\documentclass{bmvc2k}

\title{Soft Sampling for Robust Object Detection}

\addauthor{Zhe Wu*}{zhewu@umiacs.umd.edu}{0}
\addauthor{Navaneeth Bodla*}{nbodla@umiacs.umd.edu}{0}
\addauthor{Bharat Singh*}{bharat@cs.umd.edu}{0}
\addauthor{Mahayar Najibi}{najibi@cs.umd.edu}{0}
\addauthor{Rama Chellappa}{rama@umiacs.umd.edu}{0}
\addauthor{Larry S. Davis}{lsd@umiacs.umd.edu}{0}

\addinstitution{
 Center for Automation Research, \\
 University of Maryland \\
 College Park, Maryland, USA \\
}

\runninghead{Wu, Bodla, Singh, Najibi, Chellappa, Davis}{Soft Sampling}

\usepackage{wrapfig}

\begin{document}

\maketitle
\begin{abstract}
We study the robustness of object detection under the presence of missing annotations. In this setting, the unlabeled object instances will be treated as background, which will generate an incorrect training signal for the detector. Interestingly, we observe that after dropping 30\% of the annotations (and labeling them as background), the performance of CNN-based object detectors like Faster-RCNN only drops by 5\% on the PASCAL VOC dataset. We provide a detailed explanation for this result. To further bridge the performance gap, we propose a simple yet effective solution, called Soft Sampling. Soft Sampling re-weights the gradients of RoIs as a function of overlap with positive instances. This ensures that the uncertain background regions are given a smaller weight compared to the hard-negatives. Extensive experiments on curated PASCAL VOC datasets demonstrate the effectiveness of the proposed Soft Sampling method at different annotation drop rates. Finally, we show that on OpenImagesV3, which is a real-world dataset with missing annotations, Soft Sampling outperforms standard detection baselines by over 3\%. It was also included in the top performing entries in the OpenImagesV4 challenge conducted during ECCV 2018.
\end{abstract}

%-------------------------------------------------------------------------
\section{Introduction}
The performance of object detection for a limited number of classes has steadily improved over the past five years. This is largely due to improved CNN architectures \cite{simonyan2014very,resnet,xie2017aggregated}, detection models \cite{ren2015faster,dai2016r,dai2017deformable} and training techniques \cite{liu2018path, peng2017megdet,singh2017analysis}. Although it is encouraging to see improvement in performance year over year for an age old problem, techniques developed for object detection require detailed bounding-box annotations for images containing multiple objects, like the COCO dataset~\cite{lin2014microsoft}. 
The annotation process becomes significantly more complex as the number of classes is increased and we are more likely to miss annotating some instances. In other words, it is practically impossible to annotate every single instance in every image for 100,000 classes. 
On the other hand, annotating a fixed number of instances for each class over the whole dataset is much more feasible. For example, we could label 500 instances for classes like ``eye'', ``nose'', ``ear'', \textit{etc}. in some images rather than annotating every instance. 
This is different from the commonly used semi-supervised setting (as in \cite{rosenberg2005semi}), where there exists a completely annotated set for a small number of classes and another set of unlabeled images and the goal is to harness more data for improved performance. Our setting is close to datasets like OpenImages V3 ~\cite{openimages}, which contain partially annotated instances for a large number of classes. 
%+++++++++++++++++++++++++++++++++++++++++++++++++++++++++++++++++++++++++++++++++++++++++++++++++
\begin{figure}
\centering{
\includegraphics[width=0.95\linewidth]{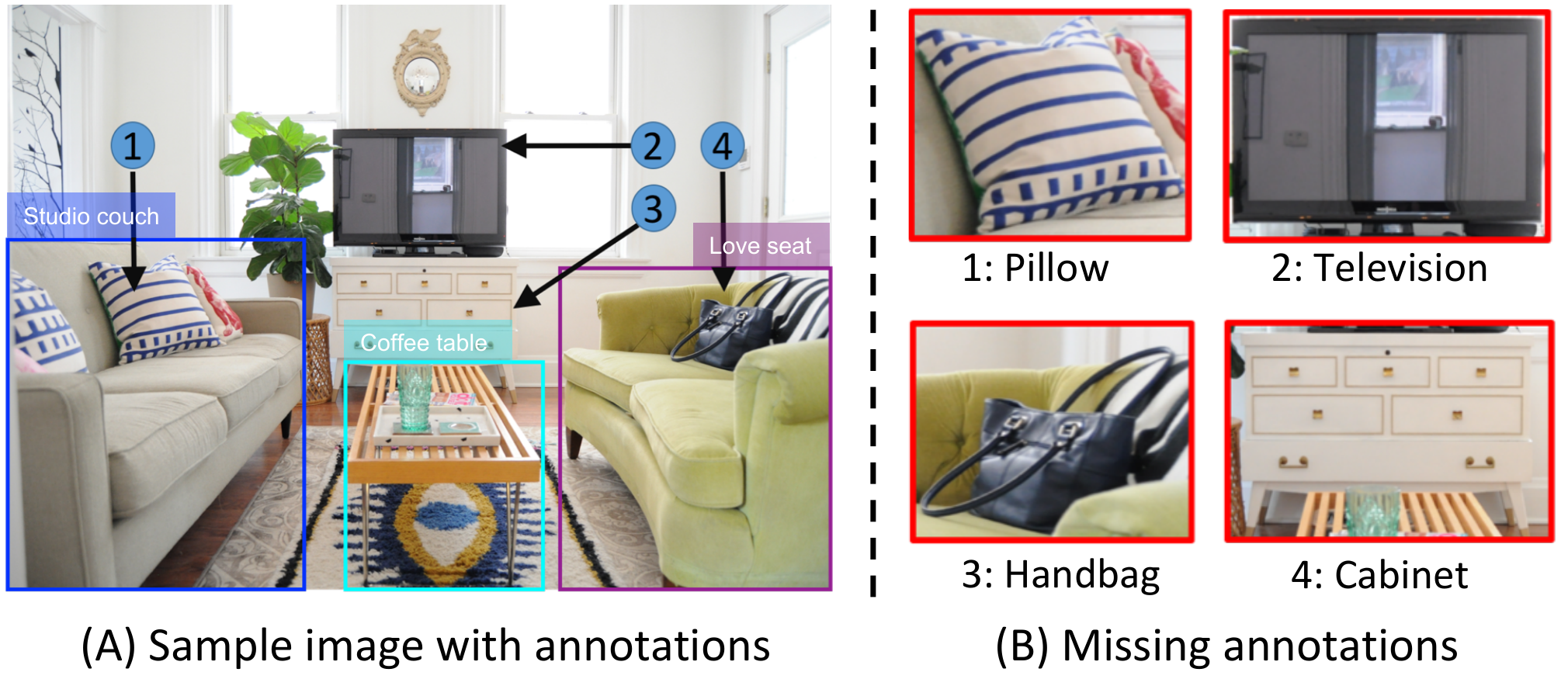}}
% \vspace{-5pt}
\caption{An example image from the training set of the Open Images dataset. (A) The ground-truth annotations are shown as colored boxes. In a scene with a rich set of objects, only three of them are annotated: love seat, studio couch and coffee table. (B) The annotations of at least four other object classes are missing which would lead to false negatives during training.}
\label{fig::teaser}
% \vspace{-5pt}
\end{figure} 
%+++++++++++++++++++++++++++++++++++++++++++++++++++++++++++++++++++++++++++++++++++++++++++++++++

Since the evaluation cycle for PASCAL VOC \cite{pascal} is faster, we curate a dataset by artificially dropping annotations present in an image to study the effect of missing annotations. On this dataset, we train a commonly used detector (Faster-RCNN \cite{ren2015faster}) and analyze how it performs when we drop different fractions of the annotations. In this setting, apart from reducing the number of positive samples, we also introduce false negatives during training the detectors. For example, in Fig.~\ref{fig::teaser}, there are atleast four objects in the image that are not annotated. Because the pillow in the image is not annotated, all the object proposals which have a significant IoU (Intersection-over-Union) with the pillow would be incorrectly marked as background. This would provide an incorrect training signal to the neural network which could deteriorate its performance by a large margin. To our surprise, this did not happen! Even after dropping 30\% of the annotations, the mAP of the detector only dropped by 5\%. To the best of our knowledge, we are the first to report that \textbf{deep learning based object detectors are fairly robust to missing annotations.} This is an important result as it simplifies the data annotation process a lot.

Although the drop in performance is not large when the annotation drop rate it low, at larger drop rates, the performance gap is still noticeable. To this end, we explore multiple training techniques which can be employed to bridge the gap between a detector which has complete supervision vs. the one trained with partial supervision. A natural solution to tackle this problem would be to use hard-example mining \cite{felzenszwalb2010object}, where negatives which do not have a minimum amount of overlap with any positive instance are completely ignored. This would reduce the false-negatives, and hence would provide cleaner gradients to the network. However, it will also ignore a significant portion of background regions in the image. We observe in our experiments that having a large pool of true negatives is also important for obtaining good performance. Based on this observation, we propose an overlap-based soft-sampling strategy, called \textbf{OSS}, which reduces the gradients as a function of overlap with positive examples to ameliorate the effect of false negatives, as illustrated in  Fig.~\ref{fig:pipe}. OSS ensures that every sampled box still participates in training, but more confident samples are given more importance. We validate the effectiveness of OSS on multiple datasets (curated and real) and show that it obtains a consistent improvement in performance.

Another approach to reduce the effect of false negatives would be to re-weight the gradients based on the confidence of a detector trained on partial annotations. Wherever the detector is more confident, we can reduce the gradients (or even change the labels). This approach would be effective in settings where the detection performance is very good. However, when the performance of the detector is low, it could generate false positives and consequently ignore regions where there are no objects. We also present results for functions which encode prior detection scores for training detectors with partial annotations. %Other reasonable techniques which did not improve performance significantly are presented in the supplementary material.
\section{Related work}
\label{gen_inst}
Large scale object detection has eluded computer vision researchers for a long time. There have been several improvements towards solving this problem but most of the work has focused on efficiency. Early efforts include training a deformable part-based model for 100,000 classes \cite{dean2013fast}. As deep learning provided a large boost in performance to visual recognition problems, detectors like YOLO-9000 \cite{redmon2017yolo9000} and R-FCN-3000 \cite{singh2017r} leveraged convolutional neural networks \cite{lecun1998gradient} for efficiently detecting a large number of classes. The main focus is again on performing efficient inference and designing a model where computational cost does not increase significantly as the number of classes is increased. Although these detectors are fast, their performance is far behind detectors trained on a limited number of classes on datasets like PASCAL and COCO. The main reason for this gap is the reduced variation in the training set, as they are primarily trained on classification datasets like ImageNet \cite{deng2009imagenet}. For example, YOLO-9000 is trained without any bounding-box supervision for most classes, due to which its mAP even on an easy detection dataset like ImageNet is 16\%. With bounding-box supervision, R-FCN-3000 was able to obtain 18\% better performance, suggesting that obtaining box-level supervision on {\em classification} data is a practical way of improving large scale object detection. Unfortunately, classification data does not include the rich context available in detection datasets like COCO. Going forward, to train detectors which can obtain human level performance, a pragmatic approach towards solving large scale detection would be to design robust detection algorithms which can learn with partial annotations on large scale detection datasets.

Training detectors with partial annotations is different from semi-supervised learning ~\cite{rosenberg2005semi,tang2016large}. In the semi-supervised setting, a clean fully annotated dataset is provided and a unlabeled dataset is leveraged for augmenting the training set. The motivation in semi-supervised learning is to discover more data to improve the performance of existing detectors. While semi-supervised learning can be leveraged for relabeling missing annotations in the partially annotated training set, when the number of classes is large, the false positive rate of current detectors is high. So, correcting regions marked as background with the correct labels is much harder than reducing the ill-effects of missing annotations. %We explore the idea of correcting labels for regions marked as background, and show that it is effective when the performance of the detector is good.

In the object detection literature, techniques like OHEM \cite{shrivastava2016training} and focal loss \cite{lin2018focal} have been proposed for sampling RoIs and reweighting the loss for anchors. However, in the missing label case, both these techniques would have adverse effects. For example, OHEM would select more background regions which are unlabelled. Similarly, in focal loss, because anchors corresponding to unannotated objects would have a high score, they would generate incorrect gradients of a high magnitude. Both these algorithms are designed to work when the dataset is fully annotated and differ from soft-sampling.
\section{Object Detection with Missing Annotations}
%formalize the problem, explain all the cases mathematically and with the help of a visual example
In object detection, an RoI (region of interest) is considered to be positive when its overlap with a positive ground-truth is greater than a pre-defined threshold, like 0.5. Otherwise, it is marked as negative. This scheme works well as long as every object in the image is correctly annotated. Let us try to understand what happens when the annotation is not correct. There are different types of incorrect annotations that are possible, like incorrect location of the bounding box, wrong label for the class or regions containing objects which are not labeled. In this work, we focus on the last case, as illustrated in Fig.~\ref{fig::teaser}. When an object present in the image is not annotated, it creates two problems. First, it reduces the total number of positive instances for the class to which it belongs. Second, a region near the instance could be incorrectly marked as background.

High performance object detectors, like Faster-RCNN, first eliminate a large amount of background regions with a region proposal network and then classify these regions using a classification network. The region proposal network places anchor boxes on the image in a sliding window fashion and selects anchors which have an overlap greater than a threshold with a ground-truth instance as positive. It randomly samples a fixed set of negative anchors (like 128) from hundreds of thousands of anchors. Therefore, the chance of these negative anchors containing a well localized object is less likely i.e, the missing annotations are less likely to create a negative effect on the network. Hence, the region proposal network only suffers from diminished training samples when we do not annotate some regions of the image. We also show in our experiments that when we train RPN with missing labels, performance of the detector does not change significantly.

However, the number of regions processed by the classification network in Faster-RCNN is significantly lower than that processed by RPN. Moreover, these regions contain a large fraction of the object instances, as RPN has a very high recall. Therefore, when label assignment is performed in the classification network, the missing annotations covered by region proposals would contribute significantly to the gradients back-propagated to the network. Therefore, the label assignment procedure seems to be not optimal when some instances are not annotated in the image. We will evaluate the robustness of different label assignment methods in the following sections.

\begin{figure}
    \center
    \includegraphics[width=0.8\linewidth]{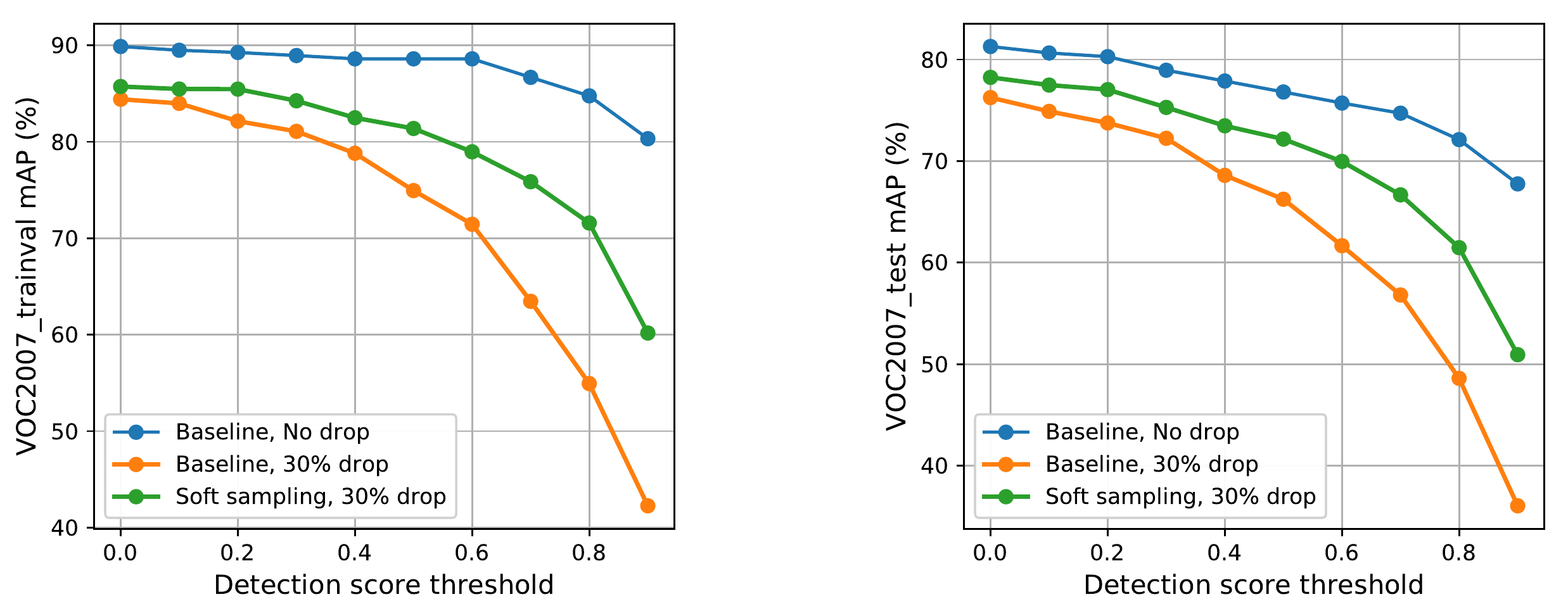}
    \caption{mAP of Faster RCNN trained with and without 30\% missing labels on the PASCAL VOC dataset at different thresholds. Left: trainval split. Right: test split.}
    \vspace{-10pt}
    \label{fig:map}
\end{figure}

\section{Evaluation of Robustness of Object Detectors to Missing Annotations}
A commonly used label assignment method is to mark any RoI (proposal) which has an overlap with a ground-truth box greater than 0.5 as positive and less than 0.5 overlap (IoU) as negative and construct a mini-batch of a fixed number of RoIs per image. We train a Faster-RCNN with this label assignment scheme and evaluate its performance when it is trained with and without missing labels. For the missing label case, we artificially drop 30\% of the ground-truth annotations per class in the PASCAL VOC dataset. We plot the mAP of the detector trained with missing labels ($\mathcal{D}_m$) and the one trained on all the annotations ($\mathcal{D}_f$) in Fig. \ref{fig:map}. We also evaluate their mAP at different detection thresholds. It is interesting to note what happens in this case. Under the evaluation criterion of mAP at 0.5 overlap, the performance gap between the two detectors is not a lot! The gap starts to increase as we increase the detection threshold. Since $\mathcal{D}_m$ is trained with missing labels, it reduces its confidence for correct samples as well, so that it is not penalized by the loss function by a large margin. Since mAP at lower thresholds is still good, it implies that the relative ranking of object instances is preserved even in the presence of missing labels but the absolute scores generated by the detector change. The same pattern holds true on the training set. Even after dropping 30\% of the samples per class, the mAP of the detector did not go below 70\%; on the contrary, it is 84\%. If the detector was overfitting to the regions which had missing labels, at least these 30\% of the regions should have been classified as background. 
This is different from the observation in the classification literature that deep neural networks are prone to overfitting \cite{zhang2016understanding}.

We interpret the results discussed above as follows: the detector assigns a very high score to correctly labeled positive regions and also assigns a very low score to the regions which do not have any object, but it still assigns a relatively high score to regions which have an object instance but are not labeled. Therefore, mean average precision, which depends on the relative ordering of scores is not affected significantly. If classification accuracy was computed, many of these detections would be marked as background due to their low score! Hence, it is important for practitioners to tune the detection threshold per class, when using detectors trained on missing labels. With that in mind, \textbf{it is possible to obtain high performance even with detectors trained on missing labels.} Current datasets like COCO have a recall of 83\% for ground truth-annotations (17\% of actual instances are missing) and large scale datasets like OpenImagesV4 have a recall of 43\% \cite{kuznetsova2018open}.  Still, the performance of detectors trained on these datasets is very good. Our ablation study systematically explains this observation. Therefore, we can significantly reduce the annotation cost as algorithms do not have a hard requirement that every instance in the image needs to be annotated. 

\section{Soft Sampling for Robust Object Detection}
%+++++++++++++++++++++++++++++++++++++++++++++++++++++++++++++++++++++++++++++++++++++++++++++++++
% \begin{figure}
%     \center
%     \includegraphics[width=1\linewidth]{imgs/foreP_ov.pdf}
%     \caption{Blue: Foreground ROI probability histogram on overlaps. Yellow: Overlap-based weighting curve on background bounding box proposals.}
%     \label{fig:fore_prob}
% \end{figure}

%+++++++++++++++++++++++++++++++++++++++++++++++++++++++++++++++++++++++++++++++++++++++++++++++++

While we showed that the performance drop for object detectors is not significant in the presence of missing labels, the drop is still noticeable. Unlabeled regions containing a positive object instance contribute to incorrect gradients back-propagated through the network. A simple technique to reduce the error would be to adopt hard example mining \cite{felzenszwalb2010object} for training detectors. In this method, any RoI which does not have a minimum overlap (like 0.1 or 0.2) with a positive instance is not included in training. This ensures to some extent that incorrect gradients from missing labels does not affect the training process. For the PASCAL dataset, in Fig.~\ref{fig:fore_prob}, we plot the probability of observing another object instance at different overlap thresholds. At high thresholds like 0.2, we can be fairly confident that incorrect gradients from missing labels will not interfere with the training process. While hard example mining solves one problem, it creates another one - it does not process the regions which are likely to be objects which are far away from the object. For example, if a bottle looks like a pedestrian, it will never participate in training which makes it a less attractive alternative to the na\"ive baseline which observed all samples.

\subsection{Overlap Based Soft-Sampling}

\begin{wrapfigure}{hR}{0.5\textwidth}
\centering
\vspace{-25pt}
\includegraphics[width=0.45\textwidth]{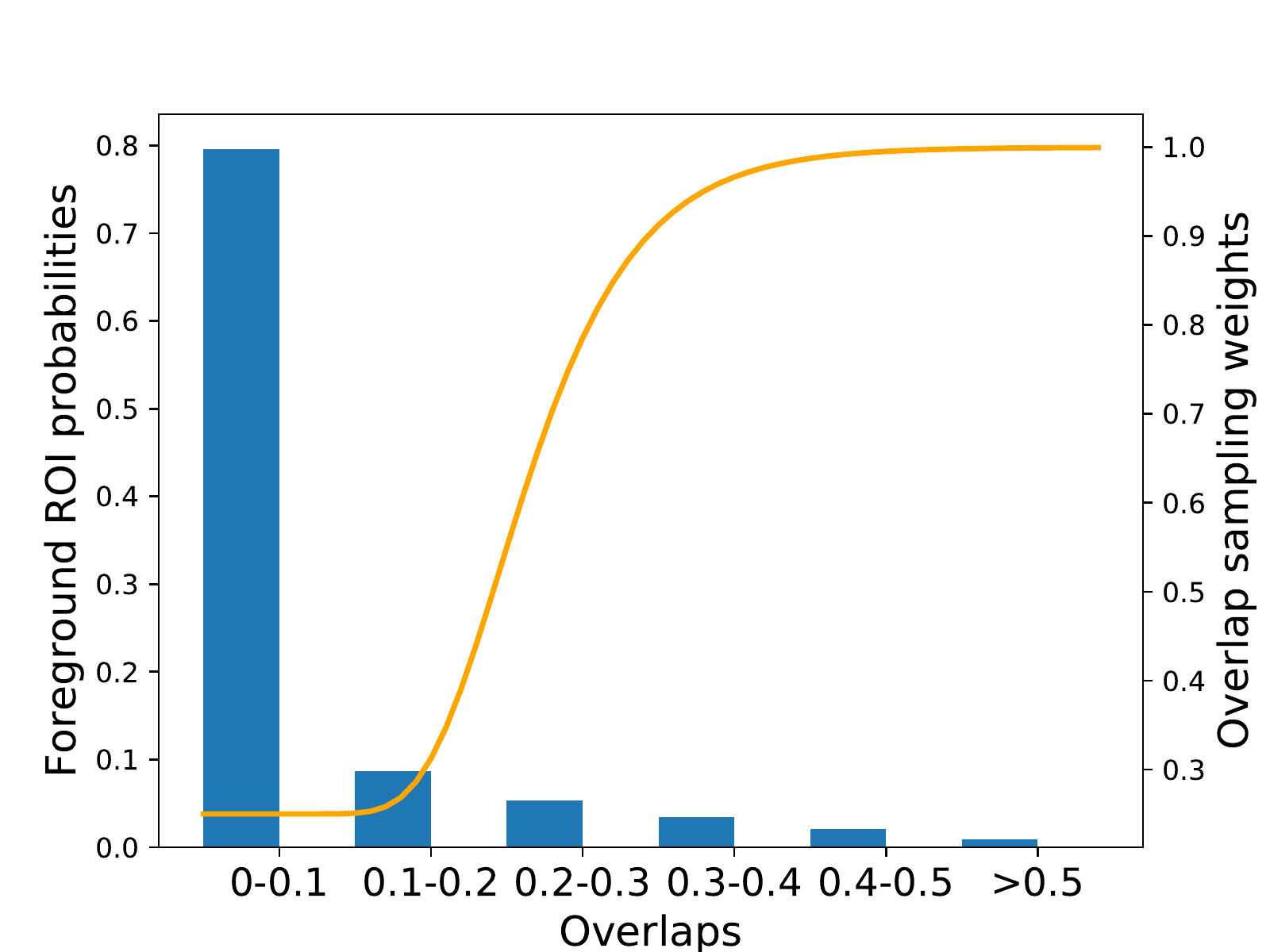}
\caption{
\label{fig:fore_prob}
Blue: Foreground ROI probability histogram on overlaps. Yellow: Overlap-based weighting curve on background bounding box proposals.}
\end{wrapfigure}

\begin{figure*}
    \center
    \includegraphics[width=1\linewidth]{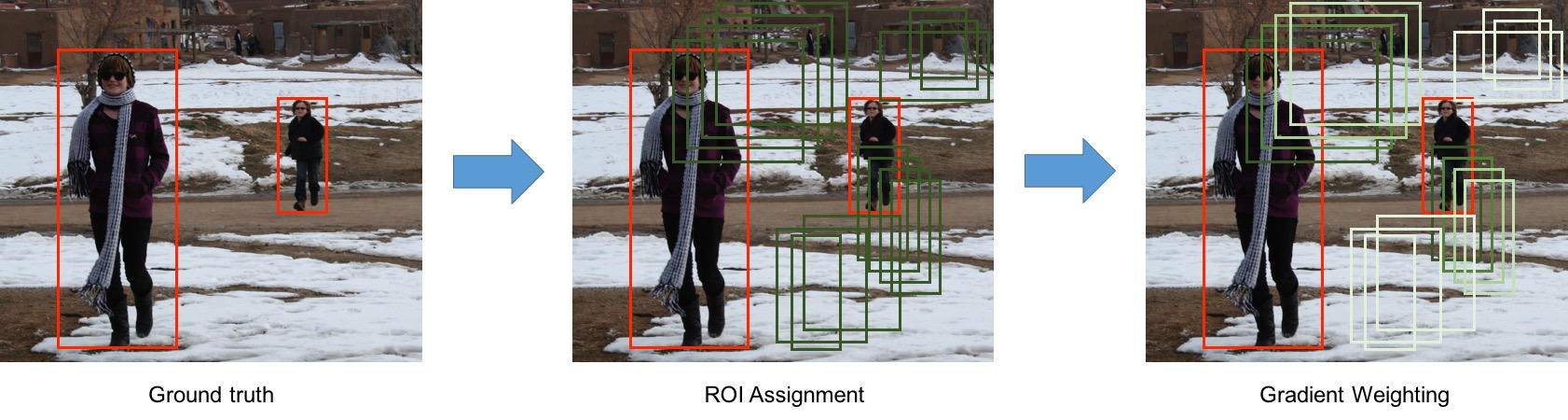}
 
\caption{Pipeline of overlap based soft-sampling. Left: two ground-truth boxes. Middle: proposals which have different overlaps with the ground-truth boxes. Right: the proposals are weighted by overlap based soft-sampling. Green and white color indicate high weight and low weight respectively.}
% \vspace{-10pt}    
    \label{fig:pipe}
\end{figure*}

Therefore, we propose to decay the gradients for each RoI as a function of its overlap with positive instances. This ensures that all positives and hard negatives will contribute significantly to the gradients. Background regions still contribute to the gradient, but with a lower weight. Soft-sampling (SS) helps us in obtaining a reasonable balance between sampling hard examples and the rest of the background. In our implementation, it is instantiated with a Gompertz function as follows,

%write this equation
$$\mathcal{G}(o) = a+(1-a)\mathrm{e}^{-b\mathrm{e}^{-co}}$$

where $a$ is approximately the weight when overlap $o = 0$ (as b is a large number, like 50), $b$ sets the displacement along the x-axis (translates the graph to the left or right), and $c$ sets the growth rate (scaling along the y axis). Gompertz function is a special form of the generalized logistic function and it grows rapidly when $o$ is small. As can be seen in Fig~\ref{fig:fore_prob}, the probability of observing another foreground RoI drops rapidly when $o$ is small. Therefore, we can rapidly increase our weights for hard negative examples using the growth rate $c$. We plot this function in Fig~\ref{fig:fore_prob}, with $a=0.25$, $b=50$, $c=20$. Our soft-sampling function assigns high weights to regions which are positives and hard negatives as we are very confident that their labels are correct. Overlap based SS (OSS) assigns a low weight to background regions as it is not sure about their labels. OSS is a simple, but effective way of assigning labels when we have missing labels and it can be easily plugged into most detectors without requiring multiple training rounds. The pipeline of OSS is shown in Fig.~\ref{fig:pipe}.

\subsection{Detection Score Based Soft-Sampling}
Another way to reduce the effect of missing labels is to apply a pre-trained detector (trained on missing samples) on the training set and obtain its detection score for each RoI. We assume that the detector would generate a high score where there are missing labels, therefore we can re-weight the gradients of RoIs which are \textbf{not} hard negatives or positives. The weighting function we use is similar to the overlap based soft-sampling and is defined as follows,

$$\mathcal{G}(s, T) = a+(1-a)\mathrm{e}^{-b\mathrm{e}^{-c(T-s)}}$$

where $T$ is the per class threshold, which is selected as the median detection score of annotated ground-truth boxes. This method would be effective if the false positive rate of the detector is low, otherwise, it would re-weight the gradients for true negatives which would adversely affect the performance of the detector. We also need to train the detector twice when using this approach.

While the above two methods would reduce the negative effect of missing labels, they would not increase our pool of positive labels. Harvesting more training data using a trained classifier is known to provide a bump in performance. This is commonly done in semi-supervised learning \cite{rosenberg2005semi}. Unfortunately, in  large scale setting, current detectors do not perform very well and their false positive rate is high. Therefore, incorrect gradients would  be generated for false positives, which would have an adverse effect on training. Therefore, we primarily focus on reducing the effect of incorrect gradients arising due to unlabeled samples instead of correcting them. 

% \subsection{Re-training with Soft Labels}
%  Recently, it was shown that we can use the predictions of a network as soft-labels and re-train it \cite{}. We also evaluate this method for re-labeling RoIs which are not hard negatives or positives. The labels in this case are defined as follows, 

% $$\mathcal{L}(x) = $$

% where blah blah is blah blah. We find this approach to be effective when the false positive rate of the detector is low.

%based on the above explaination, explain the baseline, hard negative mining as to why it is good, show HNM with a visual example

%explain why HNM is bad? it reduces the number of negatives, show some stats

%describe the soft-sampling function

%how to use detection scores with soft-sampling?

%changing labels
\section{Experiments}
We evaluate our method on two datasets, PASCAL VOC ~\cite{pascal} and Open Images V3 ~\cite{openimages}. As the evaluation cycle is faster on the PASCAL dataset, we curate multiple versions of it with different percentages of missing annotations. The PASCAL dataset also helps us to evaluate if we are able to bridge the gap between algorithms trained with partial and complete annotations. Open Images V3 inherently contains partial annotations, so it is suitable for evaluating our method. More details about training and evaluation would be described later in this section.
%To evaluate our method, we perform experiments on two datasets, PASCAL VOC ~\cite{pascal} and Open Images datasets V3 ~\cite{openimages}. Since our method proposes a solution for large scale object detection, Open Images is suitable for evaluating and analyzing the proposed method. Data collection for large scale object detection is always a challenge due to very large number of object classes and number of objects per image. Hence in practice, it is not possible to annotate all the objects in an image. Open Images V3 annotates at least one object per class per image which makes it a suitable dataset to show the performance in the weakly supervised setting. 

\subsection{PASCAL VOC}
%More training data helps in training a better object detector but the trade off between weekly annotating a dataset and performance drop is not very well studied. Large scale object detection primarily deals with missing labels and it is not known how much of a performance improvement can be expected by annotating more training data. Since 

In this section, we evaluate if it is possible to obtain high performance detectors without annotating every object instance in the training dataset. PASCAL VOC is a fully annotated dataset where popular object detectors such as Faster-RCNN \cite{ren2015faster} obtain a high mAP score. Therefore, we remove $50$\%, $40$\%, $30$\% and $0$\% of the ground-truth annotations from the training data and compare the performance of Faster-RCNN trained under these conditions. We ensure that the ground-truth boxes are removed uniformly across all the classes. Images that have only one ground-truth bounding box in an image are not dropped to ensure at least one annotation per image. To establish the performance bounds of this simulated dataset, we design the following two experiments. 

%Images that have only one bounding boxes are highly likely to be annotated in the training set and hence we don't drop them. 
%Annotating every instance of an object in an image is a tedious task. Annotating every instance, like all the chairs around a dining table or every person in a busy street seems like a pointless exercise. But, if an image contains just one car, it is more likely to be annotated and hence we attempt to simulate this setup of data collection. %+++++++++++++++++++++++++++++++++++++++++++++++++++++++++++++++++++++++++++++++++++++++++++++++++
% \begin{wrapfigure}{R}{0.5\textwidth}
% \centering
% \includegraphics[width=0.5\textwidth]{imgs/lb_ub.jpg}
% \caption{
% \label{fig:lb_ub}
% % \vspace{-10pt}
% Example illustrating negative sampling for upper bound and baseline experiments with missing labels. In A, only two of the four chairs are annotated. For baseline experiments the proposals shown in B will be labeled as background leading to false negatives whereas for the upper bound experiments, these proposals are ignored.}
% \vspace{-10pt}
% \end{wrapfigure}

%+++++++++++++++++++++++++++++++++++++++++++++++++++++++++++++++++++++++++++++++++++++++++++++++++

\noindent \textbf{Upper bound:} In this experiment, along with dropping the ground-truth annotations, we also ignore the proposals matching the dropped ground-truth bounding boxes. This is to ensure that no incorrect gradients are back-propagated during training. %For example, if an image contains $4$ chairs, if the bounding box annotations for $2$ of them are dropped, it is possible that the proposals around these dropped bounding boxes would be wrongly labeled as background. To avoid this, when we remove $2$ chairs, we also ignore the proposals that would otherwise be labeled as background (i.e, false negatives are avoided). This ensures that background and foreground classification is accurate, and therefore is the best possible performance that can be achieved {\em with missing labels}.

\noindent \textbf{Baseline:} In this experiment, we drop a percentage of the ground-truth bounding boxes and assign the proposals as foreground or background based on the remaining ground-truth bounding boxes. 

%\begin{figure*}[t]
%    \center
%    \includegraphics[width=1\linewidth]{imgs/lb_ub2.jpg}
%    \caption{Example illustrating negative sampling for upper bound and baseline experiments with missing labels. In A, only two of the four chairs are annotated. For baseline experiments the proposals shown in B will be labeled as background leading to false negatives whereas for the upper bound experiments, these proposals are ignored.}
%    \label{fig:lb_ub}
%\end{figure*}

The upper bound is the ideal scenario with the best possible performance, while the baseline is a lower bound. That is, if standard object detectors are trained as is, without any consideration to missing labels, at least baseline performance will be obtained. %An illustration of the baseline and upper bound setups are shown in Figure \ref{fig:lb_ub}.

\noindent \textbf{Sampling Hard Negatives} Along with the above two experiments, we design another baseline for reducing the effect of missing labels - sampling hard negatives. For this experiment, we sample only those negatives which have at least an overlap of $0.1/0.2$ with the existing ground-truth boxes. We expect this setup to perform better than the baseline but not necessarily as good as the upper bound (as it ignores a significant fraction of the background image). When the number of proposals is less than the batch size, many examples in the top ranked proposals (typically 300 or 500) may not have an overlap of 0.1/0.2. Therefore, we first generate more proposals for this baseline (6000) and then select the top 300 proposals which have an overlap greater than 0.1/0.2 with annotated ground-truth boxes.

\noindent \textbf{Training details and Discussion:} We perform all our experiments using Fast-RCNN with ResNet-50 \cite{resnet} backbone and proposals extracted from a pre-trained RPN. Deformable convolutions \cite{dai2017deformable} are used in the conv5 layers. For post-processing, we use Soft-NMS \cite{bodla2017soft}. We trained the RPN branch in Faster-RCNN after dropping $30$\% of the ground-truth instances and observed that there was no difference in the performance of the detector (the classification branch was trained with all the ground-truth instances). 

%\begin{figure*}[t]
%    \center
%    \includegraphics[width=0.9\linewidth]{imgs/ablation.pdf}
%    \caption{Ablation study on different parameters of the overlap based weighting function}
%    \label{fig:abl}
%\end{figure*}

\begin{table}[bt]
\begin{center}
%\small

\begin{tabular}{|c|c|c|c|c|}
\hline
  Method   & no drop & 30 \% & 40 \% & 50 \% \\
\hline\hline
Baseline & 81.28 & 76.75  & 74.62 & 69.11 \\
\hline
Hard Negative & - & 74.68  & 73.99 & 71.41 \\
\hline
Soft Sampling (Ov) & - & 78.24  & 76.78 & 73.5 \\
\hline
Soft Sampling (Sc)  & - & 77.99  & 75.91 & 72.55 \\
% \hline
% Score Weight + Soft labels & - & 78.54  & 77.43 & 74.04 \\
\hline
Upper bound & - & 79.26  & 77.5 & 74.54 \\
\hline
\end{tabular}
\end{center}
\caption{mAP@0.5 PASCAL VOC 2007 test}    
\label{tab:pascal_drop}
% \vspace{-10pt}
\end{table}

The performance of OSS (overlap based soft-sampling) when compared to the baselines is shown in Table \ref{tab:pascal_drop}. Note that the baseline performance drops by  $\sim 5$\% after dropping $30$\% annotations ($81.28$\% to $76.75$\%). However, the drop in upper bound setup is only $2$\% (from $81.28$\% to $79.26$\%). Ideally, a detector that takes missing labels into account should bridge the $3$\% gap between the baseline and the upper bound mAPs (from $76.75$\% to $79.26$\%). Using OSS alone, we obtain an mAP of $78.24$\% which is $1.5$\% better than the baseline (at $30$\% drop) and $1$\% less than the upper bound. 

In the extreme case of dropping $50$\% of annotations, we observe a severe drop of $12$\% for the baseline and $7$\% for upper bound. Therefore, in this case, the maximum score of improvement is at $5$\%. OSS recovers $4$\% of this gap. Note that as the percentage of drop increases, the gap between baseline and upper bound increases, but OSS consistently performs very close to the upper bound with just $1$\% difference. Interestingly, hard negative mining which is expected to perform at least as good as the baseline, performs slightly worse for smaller percentage of drops such as $30$\% and $40$\%. However, for a $50$\% drop in annotations, it performs better than the baseline by $2$\% but is still less than OSS. In Table \ref{tab:pascal_drop} we also present some additional results using score weighting.% and score weighting with soft labels. 
%Both these strategies 
Score weighting performs as good as the overlap weight alone, however we do not observe significant gains by combining these different strategies. One reason for this could be that there is a correlation between the overlaps and scores. Hence, when overlap and score weights are used together, one of them is likely to nullify the effect of the other. Since OSS does not require re-training the detector, it is simpler to plug into existing object detectors.

\subsection{Open Images}
Open Images V3 \cite{openimages} is a large scale object detection dataset. It is 10 times larger than COCO and spans $600$ object classes ($545$ classes considered trainable). The dataset is split into a training set (1,593,853 images), a validation set (41,620 images), and a test set (125,436 images). %To facilitate the object detection task,  $3,709,509$ bounding boxes have been provided in the training set, $204,621$ in validation and $625,282$ in the test set. Because of the large scale, the train split is annotated in a fashion that if an image contains more than one instance of a category, only one of them is annotated. While in validation and test splits, complete annotations for all instances and all categories are provided. There are about $2$ boxes per image in the training set. The Open Images evaluation metric computes interpolated average precision (AP) for each class and averages it among all classes (mAP). It is similar as Pascal VOC 2007 metric, except that in Open Images there are ground-truth boxes labeled as "group-of". When evaluating detection results, all "group-of" boxes will be ignored. 
We select a subset of 50 categories from Open Images for our experiments due to the large size of the dataset. Since missing annotations is the main challenge, we focus on categories that have larger number of instances per image when selecting the subset. For this, we sort categories by the number of instances per image in the test set, which is fully annotated. Then we select categories which have more than 200 instances in the train split and more than 100 instances in test split. We call this subset as Open50. There are $35,106$ images in the train split and $5,287$ images in test split. On average, for each category there are $2.52$ instances per image in the Open50 test set. We follow the official OpenImages evaluation metric for evaluating performance.

We use the same experimental setting as in the PASCAL VOC experiments and report the performance of Baseline, Hard Negative Sampling and OSS. Since the dataset is not fully annotated, we can not evaluate the upper bound method. Besides, due to the low mAP $(42.57\%)$ of the baseline, it is not meaningful to apply the detection score based soft-sampling method on this dataset. At $0.5$ IoU, the Baseline method and Hard Negative Sampling obtain $42.57\%$ and $40.06\%$ mAP, respectively, while the proposed Overlap Soft Sampling achieves $45.92\%$ mAP. The results demonstrate the effectiveness of OSS, which outperforms the baseline by over $3\%$. Hard negative sampling is even inferior to the baseline, implying that it is important to sample background regions in the image.

%+++++++++++++++++++++++++++++++++++++++++++++++++++++++++++++++++++++++++++++++++++++++++++++++++
% \begin{table}[h]
% \begin{center}
% %\small
% \scriptsize
% \begin{tabular}{|c|c|}
% \hline
%    Method  & AP @ 0.5  \\
% \hline
% \hline
% Baseline & 42.57 \\
% \hline
% Hard Negative & 40.06 \\
% \hline
% Overlap Soft Sampling & 45.92 \\
% \hline
% \end{tabular}
% \end{center}
% \caption{AP@0.5 Open50}    
% \label{tab:open_images}
% \end{table}

Soft-Sampling was also employed in the OpenImagesV4 object detection challenge held in ECCV 2018. The entry by Baidu \cite{gao2018solution} which used Soft-Sampling obtained a score of 62.2 on the public leaderboard and was ranked $2^{nd}$. On the private leaderboard, it obtained 58.6 and was ranked $3^{rd}$. The winning entry was better by 0.04 points. The challenge had more than 450 submissions. For fast prototyping, the team created a subset of 100,000 images from the OpenImagesV4 training set. 95,000 images were used for training and 5,000 were used for testing. In ablation testing, it was observed that Soft-Sampling improved the mAP from 36\% to 37.2\% on this test set, leading to an improvement of 1.2\%. The final ensemble included models trained with Soft-Sampling. 

%\subsection{Qualitative results and analysis}
%We show a few qualitative results in Fig \ref{fig:qual} for the baseline detector when it is trained using all the samples and when it is trained with missing annotations. These are shown in the left and in the center. The results for soft sampling when it is trained with missing annotations is shown in the right. The detection scores fall significantly for the baseline when it is trained with missing annotations. Note that it is still able to localize the regions correctly, although the scores are lower. Soft-sampling has significantly better detection scores compared to the baseline, although there is a noticable drop in scores here as well.

\subsection{Qualitative results and analysis}
\begin{figure*}[h]
    \center
    \includegraphics[width=1\linewidth]{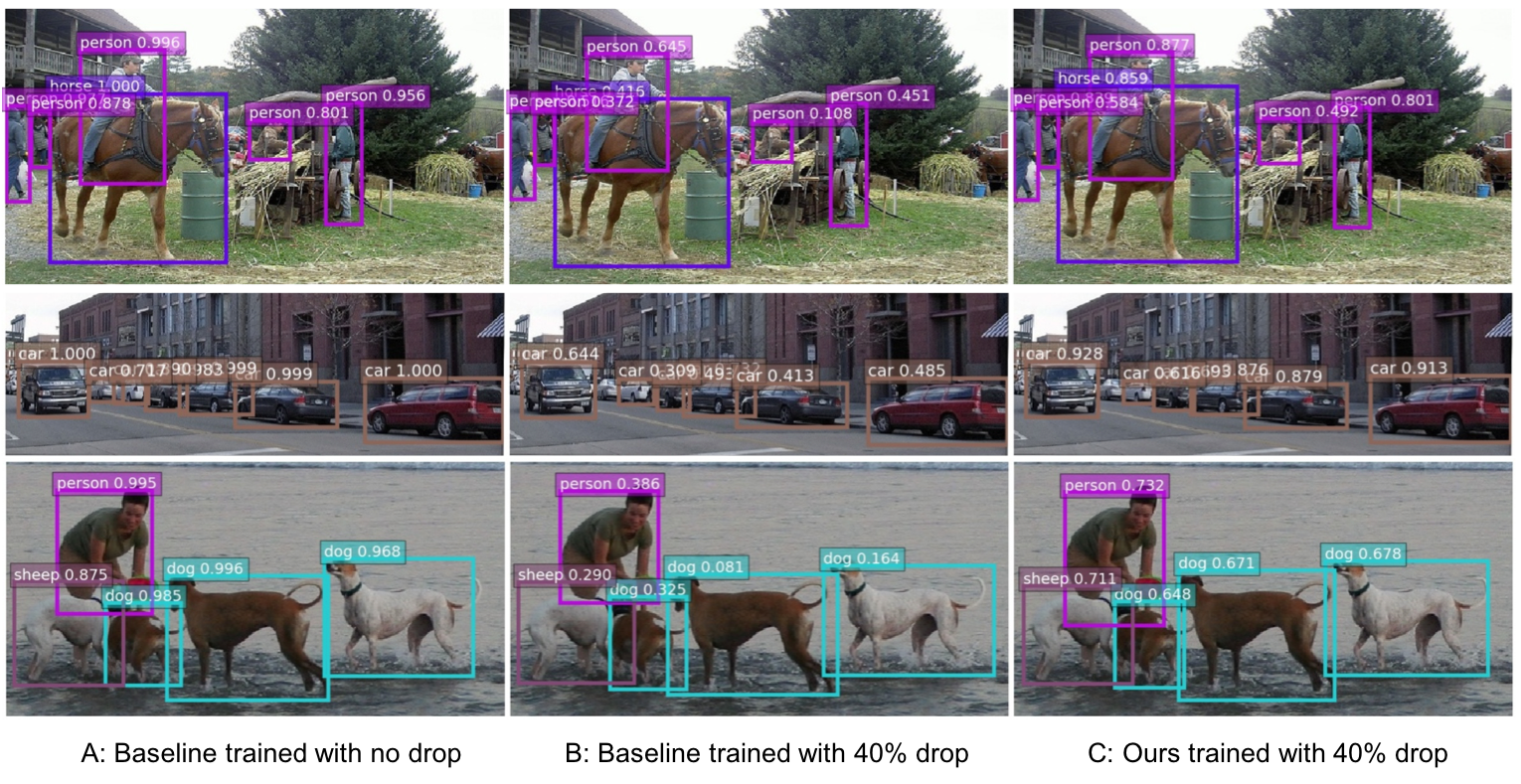}
    \caption{Visualization of detection scores. Left to right: Baseline, Baseline trained with 40\% missing annotations, Overlap based Soft Sampling trained with 40\% missing annotations. Notice how scores drop when the detector is trained with missing annotations, however, localization is still pretty good, even for the baseline detector!}
    \label{fig:qual}
\end{figure*}

We show a few qualitative results in Fig \ref{fig:qual} for the baseline detector when it is trained using all the samples and when it is trained with missing annotations. These are shown in the left and in the center. The results for soft sampling when it is trained with missing annotations is shown in the right. The detection scores fall significantly for the baseline when it is trained with missing annotations. Note that it is still able to localize the regions correctly, although the scores are lower. Soft-sampling has significantly better detection scores compared to the baseline, although there is a noticable drop in scores here as well.

\section{Conclusion}
We evaluated the robustness of modern object detection under the presence of missing annotations. Our results demonstrate that the performance gap between detectors trained on datasets which are fully annotated vs. those where a significant portion is not annotated is not as large as one would expect. This gap can be further bridged with a simple overlap based gradient weighting function to reduce the ill effects of missing labels during training. Based on our findings, it should become easier to collect large scale object detection datasets. 

\bibliography{egbib}
\end{document}